\documentclass{vgtc}                          

\graphicspath{{figures/}{pictures/}{images/}{./}} 

\usepackage{times}                     

\usepackage{tabu}                      
\usepackage{booktabs}                  
\usepackage{lipsum}                    
\usepackage{mwe}                       
\usepackage{algorithm}
\usepackage{algorithmic}
\usepackage[percent]{overpic}
\usepackage{amsmath}
\usepackage{boldline}
\usepackage{bbding}
\usepackage{newfloat}
\usepackage{listings}
\usepackage{mathptmx}                  

\onlineid{1222}

\vgtccategory{Research}

\vgtcinsertpkg

\title{CompoNeRF: Text-guided Multi-object Compositional NeRF with Editable 3D
Scene Layout}

\author{Haotian Bai\thanks{e-mail: haotianwhite@outlook.com}\\ %
        \scriptsize HKUST(GZ) %
\and Yuanhuiyi Lyu\thanks{e-mail: yuanhuiyilv@hkust-gz.edu.cn}\\ %
     \scriptsize HKUST(GZ) %
\and Lutao Jiang\thanks{e-mail:  jianglutao98@gmail.com}\\ %
     \scriptsize HKUST(GZ) %
\and Sijia Li\\
\scriptsize OPPO\\
\and Haonan Lu\\
\scriptsize OPPO\\
\and Xiaodong Lin\\
\scriptsize OPPO\\
\and Lin Wang\thanks{Corresponding author, email:  linwang@ust.hk}\\ %
     \parbox{1.4in}{\scriptsize \centering HKUST(GZ) \\ HKUST}}
\teaser{
  \centering
      \vspace{-12pt}
  {Project homepage: \url{https://vlislab22.github.io/componerf/}}
\includegraphics[width=0.9\linewidth]{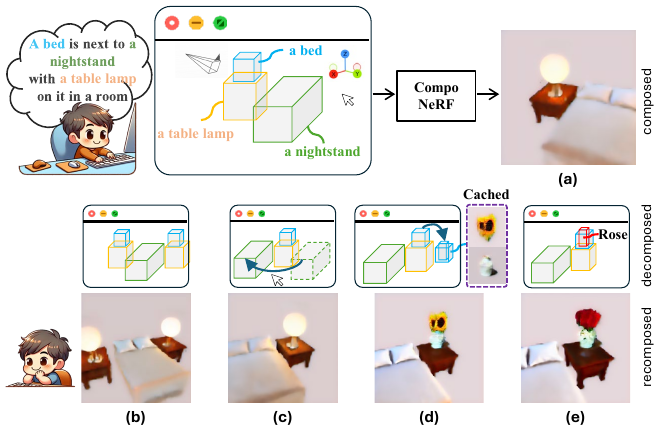}
  \caption{We introduce CompoNeRF, a novel framework that synthesizes coherent multi-object scenes by integrating textual descriptions and layouts. CompoNeRF allows for individual NeRFs, each denoted by a unique prompt color, to be \underline{composed}, \underline{decomposed}, and \underline{recomposed} with ease, streamlining the construction of complex scenes from cached models after decomposition. \textbf{(a)} displays the composed results. \textbf{(b), (c), (d), (e)} are recomposition results after manipulation demos shown above, including duplication, transformation, loading decomposed NeRFs, and semantic editing conducted separately. }
  \label{fig:teaser}
}

\abstract{
Text-to-3D form plays
a crucial role in creating editable 3D scenes for AR/VR.  
Recent advances have shown promise in merging neural radiance fields (NeRFs) with pre-trained diffusion models for text-to-3D object generation. 
However, one enduring challenge is their inadequate capability to accurately parse and regenerate consistent \textbf{multi-object} environments.
Specifically, these models encounter difficulties in accurately representing quantity and style prompted by multi-object texts, often resulting in a collapse of the rendering fidelity that fails to match the semantic intricacies. Moreover, amalgamating these elements into a coherent 3D scene is a substantial challenge, stemming from generic distribution inherent in diffusion models.
To tackle the issue of 'guidance collapse' and further enhance scene consistency, we propose a novel framework, dubbed \textbf{CompoNeRF}, 
by integrating an editable 3D scene layout with object-specific and scene-wide guidance mechanisms.
It initiates by interpreting a complex text into the layout populated with multiple NeRFs, each paired with a corresponding subtext prompt for precise object depiction. 
Next, a tailored composition module seamlessly blends these NeRFs, promoting consistency, while the dual-level text guidance reduces ambiguity and boosts accuracy. 
Noticeably, our composition design permits decomposition. 
This enables flexible scene editing and recomposition into new scenes based on the edited layout or text prompts. 
Utilizing the open-source Stable Diffusion model, CompoNeRF generates multi-object scenes with high fidelity.  Remarkably, our framework achieves up to a \textbf{54\%} improvement by the multi-view CLIP score metric.  
Our user study indicates that our method has significantly improved semantic accuracy, multi-view consistency, and individual recognizability for multi-object scene generation.

} 


\begin{document}


\firstsection{Introduction}

\maketitle
Consider the last time you wanted to relive an imaginary virtual scene through VR/AR devices. As illustrated in \cref{fig:teaser}, picture a tranquil bedroom where the day comes to a serene close. Envision a bed adorned with soft pillows, a bedside table, and a lamp casting a soothing light. Our mental imagery often includes a variety of objects and their interplay. How do we then create these scenes from descriptions, such as text prompts, and convert them into cohesive, editable environments within virtual reality?

Recent advances in text-to-image generation have been driven by the integration of vision-language pre-trained models~\cite{radford2021learning,li2022blip} with diffusion processes~\cite{ho2020denoising,nichol2021improved,rombach2022high}, leading to impressive outcomes. Pioneering text-to-3D approaches~\cite{jain2022zero,sanghi2022clip,hong2022avatarclip,JunGao2022GET3DAG,mohammad2022clip,lee2022understanding,xu2022dream3d} have built upon these successes, employing these robust vision-language models to enrich 3D generative models with the structured understanding provided by Neural Radiance Fields (NeRFs)~\cite{mildenhall2020nerf,mip-nerf,muller2022instant}. This synergy~\cite{poole2022dreamfusion,lin2022magic3d,metzer2022latent,wang2022score} facilitates the creation of 3D models which, when rendered from different views, cohere with the learned text-to-image diffusion model distribution, opening new avenues for the direct synthesis of 3D content from textual descriptions. 

\begin{figure}[t]
    \centering
    \includegraphics[width=\linewidth]{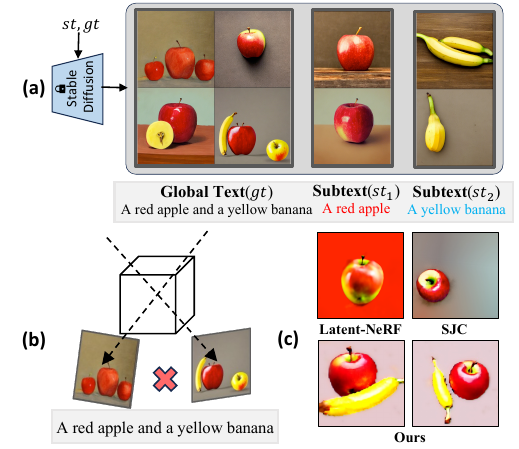}
    \vspace{-15pt}
    \caption{
    \textbf{The guidance collapse issue}.  
\textbf{(a)} Generation of the multi-object scene involves utilizing the frozen Stable Diffusion. \textbf{(b)} Instances of guidance collapse are observed when using the global text directly. \textbf{(c)} Comparison of rendering results . 
    }
    \label{fig:intro}
\end{figure}
Textual descriptions can be vague and open to interpretation. Converting these prompts into visual images, particularly for complex scenes with numerous objects, is not a simple task. 
Specifically, diffusion models like Stable Diffusion~\cite{rombach2022high} have undergone extensive training using large-scale text-image datasets~\cite{schuhmann2022laion}. Despite this, they frequently encounter difficulties when dealing with multiple object texts, particularly when those objects have a limited presence in the training data. 
As a result, this process often results in images that either leave out certain objects or depict them incorrectly. \cref{fig:intro}(a) demonstrates how even in a simple scenario involving just two objects, Stable Diffusion may not consistently preserve the integrity of the scene.
In detail, it might overlook an apple or banana or render them with inaccurate colors.
Specifically, it may neglect to include an apple or banana, or it might portray them in incorrect colors. This problem, referred to as 'guidance collapse' as illustrated in (b), is especially problematic when rendering scenes involving multiple objects based on text prompts. As shown in (c), even advanced models like Latent-NeRF~\cite{metzer2022latent} and SJC~\cite{wang2022score} struggle to accurately generate the configurations described in texts involving multiple objects. This limitation significantly hampers their capability to construct 3D scenes derived from descriptive prompts.

This raises a compelling question: \textit{Can we design a model using diffusion models armed with a generic distribution, not only identify and recreate individual elements in multi-object texts but also amalgamate them into a coherent 3D scene?}
\begin{figure}[t!]
    \centering
    \includegraphics[width=0.8\linewidth]{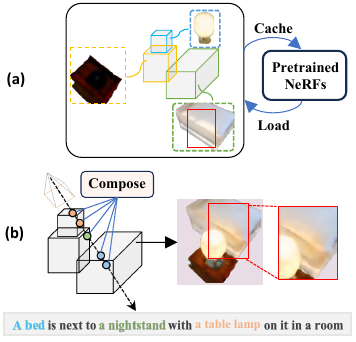}
    \vspace{-10pt}
    \caption{\textbf{(a)} CompoNeRF supports cashing and loading to facilitate NeRF composition.
    \textbf{(b)} The composition module composites multiple NeRFs for coherent scenes. Its enhanced effect is accentuated by the red boxes, showcasing superior scene coherency. 
    }
    \label{fig:compo_intro}
\end{figure}
In this paper, we present \textbf{CompoNeRF}, a compositional NeRF framework that interprets multi-object text prompts as editable 3D scene layouts with granular text prompts.
The procedure, illustrated in \cref{fig:compo_intro}(a), begins by identifying individual objects from the textual description and positioning them within customizable 3D bounding boxes. Each box is supported by a distinct NeRF and a subtext label. Thus, as shown at \cref{fig:teaser}, CompoNeRF is designed to accommodate alterations, allowing for manipulations in the layout—like moving, scaling, or removal, as well as loading decomposed nodes and direct text edition.
As depicted in \cref{fig:compo_intro}(b), our composition module guarantees that the overall scene is not merely a static collection but an orchestrated assembly. For example, after composition, the lamp illuminates the bed, creating a lively scene with such interaction among objects.

CompoNeRF distinguishes itself with three core capabilities: it \textit{composes} multi-object scenes from textual prompts, \textit{decomposes} by archiving each NeRF for subsequent utilization, and \textit{recomposes} by employing this curated content gallery to rapidly generate elaborate 3D scenes, thereby streamlining the 3D content creation workflow. 
Next, we 
employ the averaged CLIP score~\cite{wang2022clip} on rendering views of 3D content against global text prompts, we quantitatively measure the alignment of our generated scenes with their textual prompts.
For a more comprehensive evaluation, we carry out a user study to measure \textbf{1)} composition correctness regarding both semantic and multi-view consistency; \textbf{2)} generation quality in terms of users' overall preferences \textbf{3)} recognizability for each component within the scenes. 
The studies demonstrate our effectiveness in producing detailed and coherent 3D scenes that accurately reflect the given text descriptions.


To encapsulate, our paper makes three key contributions:  
(\textbf{I})  
We address the ‘guidance collapse’ problem in creating multi-object 3D scenes. Our innovative use of editable 3D layouts coupled with multiple localized NeRFs allows for precise direction over individual object representations. Moreover, these localized NeRF models are designed to be storable and reusable, enhancing efficiency in scene composition. 
(\textbf{II}) We introduce a composition module 
enables fine-tuning of the rendering process and text-based guidance, ensuring both the distinctiveness of individual objects and the holistic integration within the scene.
(\textbf{III}) 
We conduct extensive evaluations of CompoNeRF’s performance in multi-object scene generation, employing both qualitative and quantitative assessment for the multi-object text-to-3D task. The rigorous testing confirms that our CompoNeRF outperforms existing models in generating multi-object scenes that closely align with textual prompts.
\section{Related Works}
\label{sec:related_works}
\noindent\textbf{Neural Rendering for 3D Modeling.}
Recent endeavors have been made to integrate 3D content with NeRF for real-time viewing and interaction in AR/VR. For instance, works such as~\cite{rojas2023rerendrealtimerenderingnerfs, li2022rtnerfrealtimeondeviceneural}have provided real-time experience using VR/AR headsets with NeRF rendering, offering immersive virtual experiences. Recent efforts~\cite{deng2022fovnerffoveatedneuralradiance} also optimize computational resources by capturing the user’s gaze for enhanced viewing and interaction.
The evolution of NeRF has elevated the capabilities of neural rendering. NeRF-based models~\cite{nerf,lindell2021autoint,muller2022instant,Liu2020NeuralSV,mip-nerf,ref-nerf, Bai2023DynamicPF} have redefined volume rendering~\cite{kajiya1984ray} through the use of coordinate-based MLPs that infer color and density from spatial and directional inputs. Their capacity to produce photo-realistic views has cemented differential volume rendering as a key component in a variety of applications, such as scene relighting~\cite{srinivasan2021nerv,zhang2021nerfactor}, dynamic scene reconstruction~\cite{gao2021dynamic,pumarola2021d,xian2021space,tretschk2021non}, and scene and avatar editing~\cite{liu2021neural,yang2021learning}, as well as surface reconstruction~\cite{azinovic2022neural, jiang2023sdf,wang2021neus}.
Typically, these approaches rely on a \textit{single} MLP to encode an entire scene, which may introduce ambiguity in differentiating between objects. Our method, in contrast, renders scenes through a composite of multiple NeRFs, each responsible for a distinct part of the scene. This composition takes into consideration the interactions between NeRFs, allowing for the existence of individual objects while maintaining overall coherence.

\noindent\textbf{Text-guided 3D Generative Models.}
To facilitate 3D asset creation in VR/AR, the integration of vision-language models like CLIP with 3D generative methods has propelled text-guided 3D generation forward. Models that harness these advancements, such as those by Jain et al.\cite{jain2022zero} and Wang et al.\cite{wang2022clip}, have excelled in aligning 3D renderings with text descriptors but often fall short in detail, limiting realism. Innovative approaches like DreamFusion~\cite{poole2022dreamfusion}, Magic3D~\cite{lin2022magic3d}, and Latent-NeRF~\cite{metzer2022latent} have sought to enhance this through text-to-image diffusion models and score distillation sampling in the latent space, with SJC~\cite{wang2022score} and DreamBooth3D~\cite{raj2023dreambooth3d} further refining the process to address distribution mismatches and enable image-based 3D generation, respectively. Points-to-3D~\cite{yu2023points} takes a novel route by utilizing 3D point clouds for guidance, whereas Fantasia3D~\cite{Chen_2023_ICCV} innovatively disentangles geometry and appearance tasks, it employs the Stable Diffusion model for learning geometry and utilizes the Physically-Based Rendering (PBR) material model~\cite{McAuley_Hill_Hoffman_Gotanda_Smits_Burley_Martinez_2012} for appearance learning. 
Moreover, several studies~\cite{liang2024luciddreamer, wang2024prolificdreamer, yu2023text, ma2024scaledreamer, li2024connecting} are dedicated to enhancing the SDS loss to provide more detailed supervision.
%
%
Departing from these singular approaches, our CompoNeRF introduces a novel approach for the creation of multi-object 3D scenes. It adopts an \textit{object-compositional} strategy, utilizing an editable 3D scene layout that conceptualizes the scene not as a singular but as a constellation of discrete NeRFs. Each NeRF is associated with its spatial 3D bounding box and a corresponding text prompt, allowing for guidance from both the global texts and their subtexts. This dual-text framework ensures that each object is not only individually delineated but also integrated into the composite scene, thereby enhancing the authenticity of the generated 3D scenes.
%

\noindent\textbf{Object-Compositional Scene Modeling.}
The creation of new scenes from individual, object-centric components represents a trend in scene generation, as evidenced by existing research~\cite{zhi2021place, yang2021learning, wu2022object, mirzaei2022laterf, xu2022discoscene, song2022towards}. These efforts typically adopt one of two approaches: semantic-based or 3D layout-based.
Semantic-based methods enhance object representations by incorporating additional semantic information, such as segmentation labels~\cite{zhi2021place}, instance masks~\cite{yang2021learning, wu2022object}, or features extracted using pre-trained vision-language models~\cite{mirzaei2022laterf}. On the other hand, 3D layout-based approaches, exemplified by NSG~\cite{ost2021neural} and its successors~\cite{song2022towards, epstein2024disentangled}, focus on spatial coordinates, using explicit 3D object placement data to guide object and scene composition.
Diverging from conventional techniques, our method innovates by utilizing decomposed, object-specific 3D layouts. This approach enables precise control over scene dynamics, encompassing both object-specific text prompt modifications~\cite{poole2022dreamfusion, lin2022magic3d} and spatial manipulation.
%
CompoNeRF's distinctive feature lies in its capability to recompose scenes by interfacing with decomposed NeRFs, thereby accelerating the creation of new scenes. In contrast to the mesh-based method in Fantasia3D, which requires considerable human effort in mesh modification and graphics engine support for editing, CompoNeRF offers a more streamlined process. Our composition module seamlessly integrates components, requiring minimal adjustments in layout or text prompts, followed by fine-tuning existing offline models to align with the global context during training.
\section{Method}
\label{sec: method}
To resolve the issue of guidance collapse, our principal strategy is to \textit{decompose the scene into reusable components and compose/recompose them into a unified and consistent one}.
This enables flexible control over the generated content with direct use of prompts and box layouts.
As illustrated in \cref{fig:teaser}, our proposed CompoNeRF confers several key benefits:
1) \textbf{Semantic Coherence}: It reliably creates 3D objects with detailed textures and global consistency, exemplified by authentic light interactions, such as reflections on the bed surface.
2) \textbf{Modularity and Reusability}: CompoNeRF functions as an ensemble of independently trained NeRF models. These can be efficiently stored and later retrieved from a cached dataset, enabling their reuse in various cases.
3) \textbf{Editability}: Our approach allows for flexible scene modification, such as interchanging the lamp for a vase filled with sunflowers or altering its scale, by simply adjusting the box dimensions for later finetuning. This feature enhances flexibility and creative possibilities.


\subsection{Preliminaries}
Defining individual object bounding boxes as \textit{local frames} and the overall scene coordinate system as the \textit{global frame}, we build the foundation of NeRF and diffusion processes.

\label{sec:background}
\noindent \textbf{3D Representation in Latent Space.}
Our methodology capitalizes on the state-of-the-art text-to-image generative model—Stable Diffusion as described by Rombach et al\cite{rombach2022high}.
We build upon the Latent-NeRF framework~\cite{metzer2022latent}, which computes latent colors for individual objects by considering their sample positions within a localized frame. Specifically, it maps a three-dimensional point in local coordinates \(\boldsymbol{x}_l = (x_l, y_l, z_l)\) to a volumetric density \(\boldsymbol{\sigma}_l\) and an associated color \(\boldsymbol{C}_l\), expressed as \((\boldsymbol{C}_l, \boldsymbol{\sigma}_l) = f_{\boldsymbol{\theta}_l}(x_l, y_l, z_l)\). Here, \(f\) represents a Multi-Layer Perceptron (MLP) characterized by parameters \(\boldsymbol{\theta}_l\).
 This NeRF-generated color is then assessed in the context of the Stable Diffusion model, using text prompts to guide NeRF toward spatially coherent inference with intricate context.
\begin{figure*}[t]
    \centering
    \includegraphics[width=0.9\linewidth]{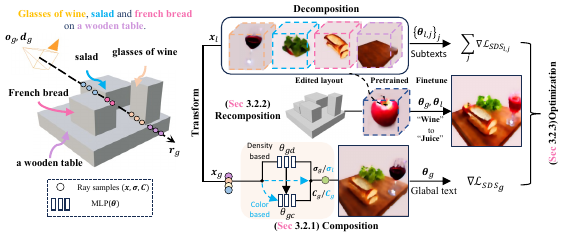}
    \caption{\textbf{Framework Overview}.
The CompoNeRF model unfolds in three stages: 1) Editing 3D scene, which initiates the process by structuring the scene with 3D boxes and textual prompts; 2) Scene rendering, which encapsulates the composition/recomposition process, facilitating the transformation of NeRFs to a global frame, ensuring cohesive scene construction. Here, we specify design choices between density-based or color-based(without refining density) composition; 3) Joint Optimization, which leverages textual directives to amplify the rendering quality of both global and local views, while also integrating revised text prompts and NeRFs for refined scene depiction.
    }
    \label{fig:framework}
\end{figure*}
\noindent \textbf{Volume Rendering with Multiple Objects.}
We extend the volume rendering process to accommodate multiple objects by assigning each a local frame, denoted as $j$, with NeRF parameters $\boldsymbol{\theta}_{l, j}$. Drawing from the foundational NeRF approach \cite{nerf}, in each local frame, we integrate the color $\boldsymbol{C}_l$ and density $\boldsymbol{\sigma}_l$ for points $\boldsymbol{x}_l$ sampled along a ray $\boldsymbol{r}_l$, emanates from the camera origin $\boldsymbol{o}_l$ in direction $\boldsymbol{d}_l$. This is formalized in the predicted color integration for $\hat{\boldsymbol{C}}_l$ as:
{\setlength\abovedisplayskip{2pt}
\setlength\belowdisplayskip{2pt}
\begin{equation}
\label{eq:volrend}
{\hat{\boldsymbol{C}}_l}({\boldsymbol{r}_l})=\sum_{k=1}^{N} T_{l, k} \left(1-\exp \left(-\sigma_{l, k} \delta_k\right) \right) {\boldsymbol{C}}_{l,k},
\end{equation}}where $T_{l, k}=\exp \left(-\sum_{j=1}^{k-1} \sigma_{l,j} \delta_j\right)$ represents the transmittance to the $k$-th of total $N$ sample, calculated exponentially over the cumulative density along $\boldsymbol{r}_l$, and $\delta_k$ is the interval between adjacent samples.
To synthesize a coherent scene, we transition from processing individual local frames to a collective global frame. Within this global context, we reconcile object attributes inferred from their individual local NeRFs for refined $\boldsymbol{\sigma}_g, \boldsymbol{C}_g$ along with $T_{g, k}$. The samples $\boldsymbol{x}_g$ are ordered based on their spatial distances from the origin $\boldsymbol{o}_g$ following the coordinate transformation. We then express the volumetric rendering of a ray $\boldsymbol{r}_g$ integrating $m$ objects within the global frame as follows:
{
\setlength\abovedisplayskip{2pt}
\setlength\belowdisplayskip{2pt}
\begin{equation}
\label{eq:multi_volrend}
{\hat{\boldsymbol{C}}_g}({\boldsymbol{r}_g})=\sum_{k=1}^{m*N} T_{g, k} \left(1-\exp \left(-\sigma_{g, k} \delta_k\right) \right) {\boldsymbol{C}}_{g,k}. 
\end{equation}}

\noindent \textbf{Score Distillation Sampling.}
To facilitate the conversion from text descriptions to 3D models, DreamFusion~\cite{poole2022dreamfusion} utilizes Score Distillation Sampling (SDS), leveraging the generative capabilities of a diffusion model, denoted as $\phi$, to guide the optimization of NeRF parameters, symbolized as $\boldsymbol{\theta}$.
Initially, SDS creates a noisy image $\boldsymbol{X}_t$ by infusing a randomly sampled noise $\epsilon$, which follows a normal distribution $\mathcal{N}(0, I)$, into a NeRF-rendered image $\boldsymbol{X}$ at a given noise level $t$.
The diffusion model $\phi$ then estimates the noise $\epsilon_\phi\left(\boldsymbol{X}_t, t, T\right)$ from this noisy image, conditioned by the noise level $t$ and an optional text prompt $T$. 
The key step in SDS involves calculating the gradient of the loss function, which measures the discrepancy between the estimated noise and the originally added noise:
{\setlength\abovedisplayskip{2pt}
\setlength\belowdisplayskip{2pt}
\begin{equation}
\label{eq:sds_loss}
\nabla_\theta \mathcal{L}_{\text{SDS}}(\boldsymbol{X}_t, T)=  w(t)\left(\epsilon_\phi\left(\boldsymbol{X}_t, t, T\right)-\epsilon\right),
\end{equation}}where $w(t)$ is a weighting function that adjusts the influence of the gradient based on the noise level. 
The gradients across all rendered views direct the update of $\boldsymbol{\theta}$, ensuring that the NeRF-generated images align with the text descriptions. Additionally, we incorporate the 'perturb and average' technique from SJC for more robust $\mathcal{L}_{\text{SDS}}$. For a comprehensive understanding of these methods, the reader is directed to the detailed explanations provided in \cite{poole2022dreamfusion,wang2022score}.

\subsection{The Proposed CompoNeRF}
\subsubsection{Composition Module}
CompoNeRF is designed to composite multiple NeRFs to reconstruct scenes featuring multiple objects, utilizing guidance from both bounding boxes and textual prompts. Within our framework, depicted in \cref{fig:framework}, the Axis-Aligned Bounding Box (AABB) ray intersection test algorithm is applied to ascertain intersections across each box in the global frame. Subsequently, we sample points \(\boldsymbol{x}_g\) within the intervals of the ray-box and project them to \(\boldsymbol{x}_l\) to deduce the corresponding color \(\boldsymbol{C}_l\) and density \(\boldsymbol{\sigma}_l\) within individual NeRF models.
These properties are processed through our composition module to infer the global color \(\boldsymbol{C}_g\) and density \(\boldsymbol{\sigma}_g\), crucial for the global rendering.
Volume rendering techniques~\cite{kajiya1984ray} are then employed to procure the rendered views for both local and global frames. We propose dual SDS losses to ensure coherence between the image outputs and their corresponding textual descriptions. Additionally, our approach facilitates recomposition by channeling samples from cached models back into local frames along with the text revision, thereby streamlining the integration.

%

\noindent\textbf{Global Composition.}
The independent optimization of each local frame may inadvertently result in a lack of global coherence within the scene. To address this, our scene composition process is designed to integrate these frames, thereby achieving a more consistent result.
Before exploring the specifics of the module, it is imperative to discuss two critical design decisions within the composition module, as depicted in \cref{fig:framework}.
Upon integrating the properties inferred from \(\boldsymbol{x}_g\) into the composition module, they are fine-tuned through gradients derived from the global SDS loss.  This process leads to a critical consideration: the necessity of refining the global density \(\boldsymbol{\sigma}_g\). There are two potential methods: \textbf{1) Density-based:} The advantage of adjusting \(\boldsymbol{\sigma}_g\) is that it can adjust geometry, thus yielding a scene more congruent with the global text prompt. 
However, this comes at the cost of potentially compromising the optimal color \(\boldsymbol{C}_g\), as calibrating \(\boldsymbol{\sigma}_g\) introduces more uncertainty for subsequent color refinement as it requires prior density features $\boldsymbol{h}$.
\textbf{2) Color-based:} Conversely, directly employing \(\boldsymbol{\sigma}_l\) mitigates this uncertainty but has less geometric control, presenting a balance to strike in the pursuit of precise scene composition.

\begin{figure}[t!]
    \centering
    \includegraphics[width=\linewidth]{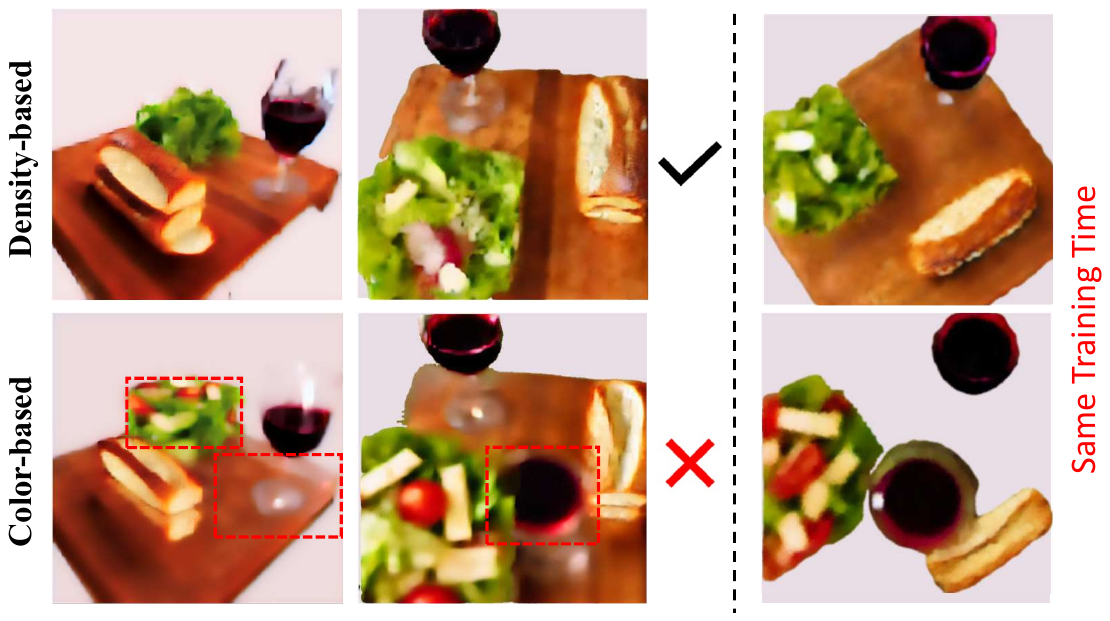}
    \caption{\textbf{Design Impact Comparison: Density vs. Color-based Methods.} The top row illustrates the density-based approach's detailed rendering and quick convergence in the 'table wine' scene. The bottom row highlights the color-based method's enhancements and its drawbacks, such as geometric and shadow inaccuracies, particularly in close-up views and slow convergence.
    }
    \label{fig:abls}
\end{figure}

After thorough experiments, exemplified in \cref{fig:abls}, we have opted for the density-based approach to refine \(\boldsymbol{\sigma}_g\)  prioritizing both \textbf{accuracy and efficiency}. The test revealed that it excels in rendering intricate details, such as enhanced wood grain textures and more naturally contoured 'salad', as accentuated by boxes. This method also demonstrated a swifter convergence rate. Conversely, while the color-based improved reflections and reduced flickering on the 'wine cup', it was plagued by issues such as sparse density, which adversely brings holes at the base of the 'cup' and the corner of the 'table'.
Furthermore, upon close examination, it becomes evident that shadow artifacts of 'wine' on the 'table' are pronounced, suggesting that its disadvantages outweigh its advantages.

\begin{figure}[t!]
    \centering
    \includegraphics[width=\linewidth]{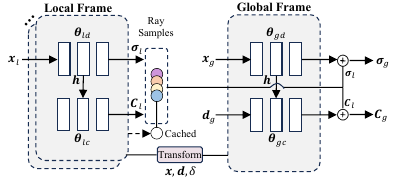}
    \caption{\textbf{Detail of Composition module}: density-based design. 
    }
    \label{fig:compo}
\end{figure}

\noindent\textbf{Network Design.}
The compositional framework of our network, as delineated in \cref{fig:compo}, is predicated on an architecture that employs a suite of MLPs, represented as \(\{\boldsymbol{\theta}_l\}_{l=1}^{m}\),  each dedicated to a distinct local frame. To harmonize \(\boldsymbol{\sigma}_l\) and \(\boldsymbol{C}_l\), we incorporate global MLPs, including density calibrator $f_{\boldsymbol{\theta}_{g_d}}$ and color calibrator $f_{\boldsymbol{\theta}_{g_c}}$.
A transformation module complements this system, tasked with maintaining the spatial coherence between the global and local frames. It governs the transformation of sampling points $\boldsymbol{x}$, ray directions $\boldsymbol{d}$, and adjacent sampling distances $\delta$. This module also orders the points $\{\boldsymbol{x}_{g,j}\}_j$ by their distance to the global camera origin $\boldsymbol{o}_g$, ensuring that each local point $\boldsymbol{x}_l$ is accurately matched with its corresponding global point $\boldsymbol{x}_g$ for subsequent volume rendering. 

The network design is:
{
\setlength\abovedisplayskip{4.5pt}
\setlength\belowdisplayskip{4.5pt}
\begin{align}
\label{eq:g_c_d}
{\boldsymbol{\sigma}_g}  &= \alpha_d f_{\boldsymbol{\theta}_{g_d}}({\boldsymbol{x}_g}) + \boldsymbol{\sigma}_l, \\
{\boldsymbol{C}_g}  &= \alpha_c f_{\boldsymbol{\theta}_{g_c}}(\boldsymbol{h}, {\boldsymbol{d}_g}) + \boldsymbol{C}_l. 
\end{align}}In contrast to the local frames, the global frame's color output $\boldsymbol{C}_g$ is inferred based on $\boldsymbol{h}$ and conditional on $\boldsymbol{d}_g$ to enable a view-dependent lighting effect.
%
%
Residual learning is leveraged here, where \(\boldsymbol{\sigma}_l, \boldsymbol{C}_l\) serve as foundational elements that support the learning of global density \(\boldsymbol{\sigma}_g\) and color \(\boldsymbol{C}_g\). The parameters \(\alpha_d, \alpha_c\) are adjustable, allowing fine-tuning of the influence that local components exert on the global outputs.
It is imperative to acknowledge that in our color-based method, density calibration is intentionally excluded to concentrate solely on the refinement of color dynamics as shown at \cref{fig:framework}. This is achieved by conditioning the process on both spatial and directional global inputs \((\boldsymbol{x}_g, \boldsymbol{d}_g)\), as demonstrated in the following equations:
\begin{align}
\setlength\abovedisplayskip{4.5pt}
\setlength\belowdisplayskip{4.5pt}
\label{eq:g_c_c}
\boldsymbol{\sigma}_g = \boldsymbol{\sigma}_l, \quad
{\boldsymbol{C}_g} = \alpha_c f_{\boldsymbol{\theta}_{g_c}}({\boldsymbol{x}_g}, {\boldsymbol{d}_g}) + \boldsymbol{C}_l.
\end{align}
The integration of extra $\boldsymbol{x}_g$ aims to facilitate a fair comparison under same inputs with the density-based. It enhances the visual appeal of effects like the wine cup's reflection, as demonstrated in \cref{fig:abls}. However, this method is not without its compromises. It tends to produce artifacts and is characterized by a slower convergence rate. Additionally, this approach limits the ability to precisely control density, subsequently impacting the intricate geometric details.

\begin{figure*}[t!]
    \centering
    \includegraphics[width=0.9\linewidth]{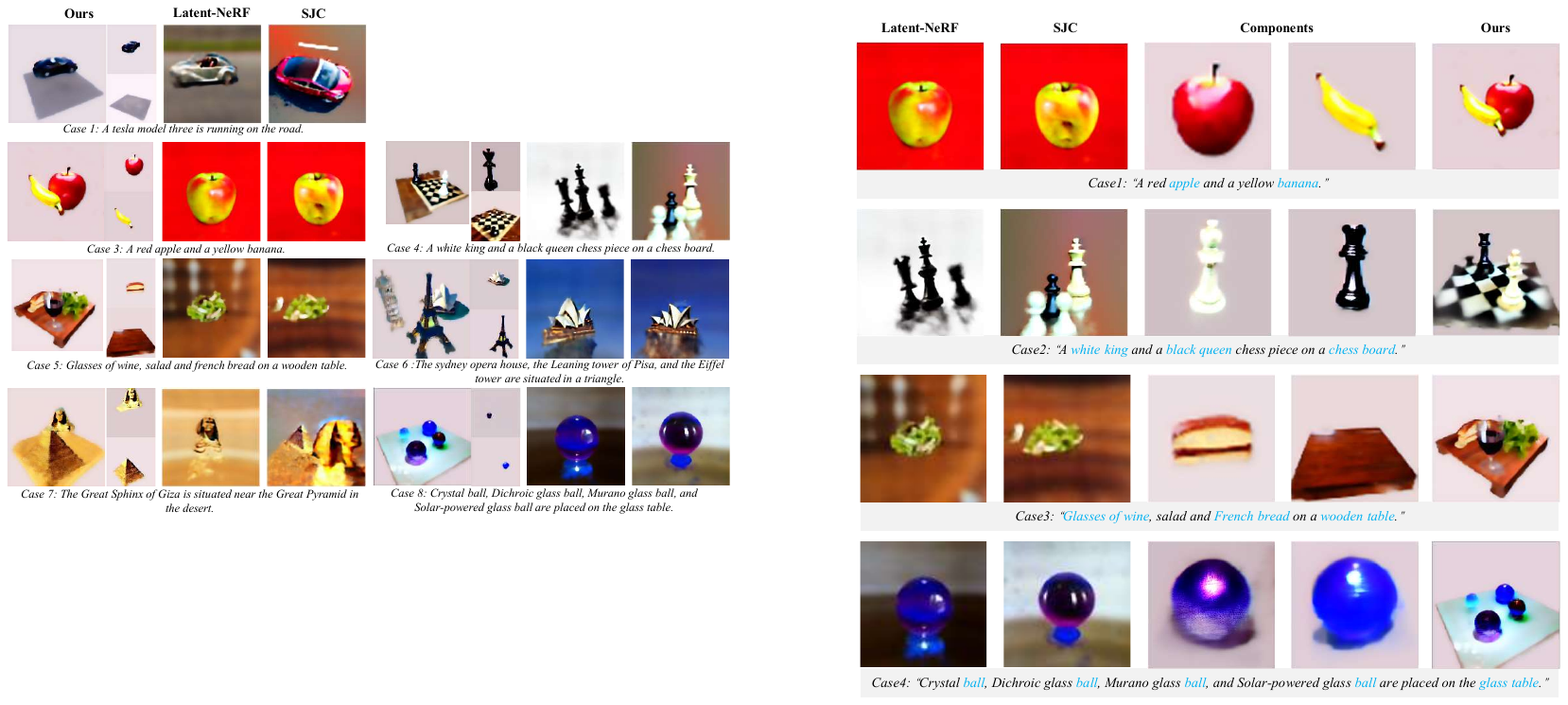}
    \caption{\textbf{Qualitative comparison with other text-to-3D methods using multi-object text prompts}. We refer readers to our \textit{suppl.} and video demos for more visual results. 
    }
    \label{fig:sota}
\end{figure*}

\subsubsection{Recomposition}
Our architecture advances scene reconstruction by providing an intuitive interface for layout manipulation.  This capability is crucial for the reconfiguration of scene elements into novel scenes, as depicted in \cref{fig:framework}. Here, the input panel allows for adjustments in the attributes of bounding boxes, such as modifying the position and scale of the 'apple' bounding box prior to composition. The refinement process further involves sampling ray-box intervals from the global frame, leading to transformed coordinates with the corresponding ray samples that are then incorporated into the pipeline, as demonstrated in \cref{fig:compo}.
Each bounding box represents a NeRF, providing the flexibility to move, scale, or replace elements as needed. CompoNeRF's capabilities also extend to textual edits, exemplified by the transformation of 'wine' into 'juice'.
Since NeRFs have been well trained, we only finetune \(\theta_g, \theta_l\) to align text prompts to promote consistency of both local and global views.
Moreover, the NeRFs once retrained within the edited scene, are also structured to be decomposable and cacheable in future scene compositions.

\begin{figure*}[t!]
    \centering
    \includegraphics[width=\linewidth]{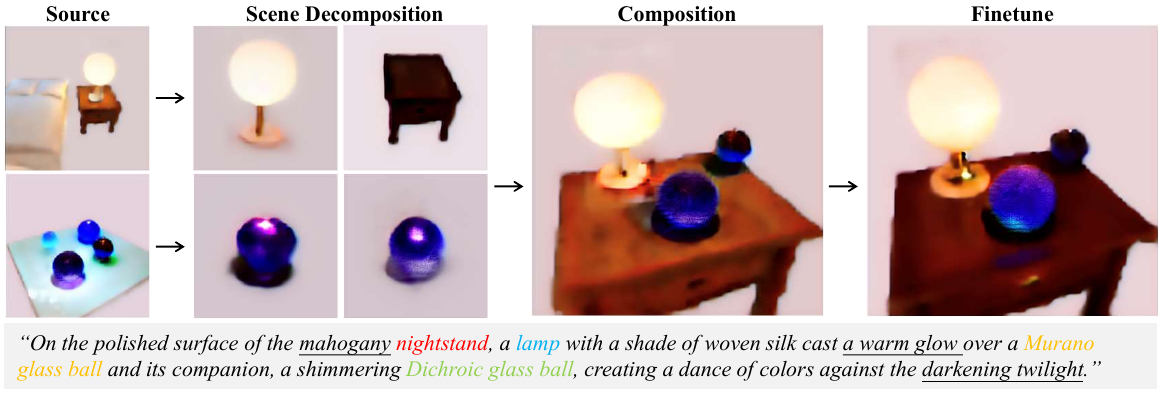}
    \caption{\textbf{The scene editing.} Demonstrated here are the stages of our recomposition, utilizing cached source scenes. Each NeRF is individually identified by colorful labels. These decomposed nodes are then positioned in the initial layout and subsequently calibrated to form the final composition. The detailed description of the ambient environment is underscored, enhancing the scene's realism.}
    \label{fig:app}
\end{figure*} 

\subsubsection{Optimization}
\label{sec:optimization}
During optimization, our method employs dual text guidance to align rendering results with both global and local textual descriptions. The optimization objective is:
{
\small
\setlength\abovedisplayskip{2pt}
\setlength\belowdisplayskip{2pt}
\begin{equation}
\label{eqn:loss_f}
\mathcal{L}= {\alpha_g}\nabla\mathcal{L}_{\text{SDS}}(\hat{\boldsymbol{X}}_{g}, T) + {\alpha_l}\sum_{j=1}^{m} \nabla\mathcal{L}_{\text{SDS}}(\hat{\boldsymbol{X}}_{l,j}, T_{l,j}) + \beta\mathcal{L}_{\text{sparse}},\nonumber
\end{equation}
}where $T$ signifies the global text prompt, while $T_{l}$ pertains to a specific object within the global context. The hyperparameters $\alpha_{g}, \alpha_{l}$, and $\beta$ modulate the respective loss weights. 
As suggested in~\cite{metzer2022latent}, we use $L_{\text{sparse}}$ included to penalize the binary entropy of local NeRFs' densities, thereby mitigating the issue of extraneous floating radiance.
Additionally, incorporating directional cues such as "front view" or "side view" into the input text, as suggested by \cite{poole2022dreamfusion,metzer2022latent} proves beneficial in specifying camera poses during the training phase, further enhancing the alignment of our generated scenes with the intended perspectives.
 We refer readers to the pseudo-code in our \textit{suppl.} for our training procedure.
\section{Experiments}
\label{sec: exp}



\subsection{Qualitative Comparison}
In Fig.~\ref{fig:sota}, we conduct qualitative comparisons of 3D assets generated using our method against Latent-NeRF and SJC, all based on the same Stable Diffusion model. Our method exhibits a remarkable ability to generate complex 3D models from a wide array of multi-object text prompts, demonstrating \textbf{superior object identity accuracy} and \textbf{enhanced context relevance and richness} compared to its counterparts.
In simpler Case 1, our CompoNeRF method adeptly generates two distinct objects: a red \emph{apple} and a yellow \emph{banana}. In stark contrast, competing methods amalgamate the features of both fruits into a singular, indistinct object.
Case 2-4, which is more intricate, showcases the capability to render a realistic scene with accurately depicted objects.
Conversely, our baselines struggle to produce even recognizable objects.
%
\begin{table}[h]
\renewcommand{\arraystretch}{1.2}
\fontsize{4pt}{4pt}
\selectfont 
\centering
\resizebox{\linewidth}{!}
{
\begin{tabular}{lcccc}
\hline
Method                   & Case 1         & Case 2         & Case 3         & Case 4         \\ 
\hlineB{1.1}
LatentNeRF               & 27.69          & 31.19          & 21.55          & 29.51          \\
SJC                      & 28.21          & 30.53          & 23.33          & 28.76          \\
\textbf{CompoNeRF (Ours)} & \textbf{33.37} & \textbf{31.45} & \textbf{36.06} & \textbf{30.98} \\ \hlineB{1.1}
\end{tabular}
}

\caption{\textbf{Quantitative comparison}. For our evaluation metric, we utilize the average of CLIP scores~\cite{parmar2023zero,zhang2023sine,wang2023imagen} across different views, which serve to assess the similarity between the generated images and the global text prompt. }
\label{tb:perclass}
\end{table}

\subsection{Quantitative Comparison}
In our study, we employ the CLIP score as the primary evaluation metric to assess the congruence between the generated 3D assets and the associated text prompts. This score, commonly used in text-to-image generation research as noted in studies~\cite{parmar2023zero,zhang2023sine,wang2023imagen}, is derived from the cosine similarity between the embeddings of the text and the image, both encoded by the CLIP model. For 3D assets, we project images from various views and calculate the CLIP score concerning the global text prompt, with the overall CLIP score being the average of these values. As detailed in \cref{tb:perclass}, our quantitative comparisons demonstrate the superior alignment of our method with the text prompt compared to the Latent-NeRF and SJC approaches in diverse scene configurations. Notably, in the challenging Case 3, our method shows a remarkable 54\% improvement. The result of overall enhancement in multi-object scenes underscores the robustness of our global calibration strategy. We refer readers to our \textit{suppl.} for additional results.
\subsection{Recomposition for more complex scenes}
To further validate CompoNeRF's performance for more complex multi-object text inputs, we assess its performance using a lengthy sentence describing the color, texture, light, and relationships between scene components.
Fig.~\ref{fig:app} showcases our refined scene renderings originating from pre-trained source scenes including the 'lamp' and the 'nightstand' from the 'bedroom' scene. Then we add two glass balls 'Murano' and 'Dichroic' from Case.4 in \cref{fig:sota}. 
We observe that a direct amalgamation of these components can manifest various artifacts at the base of the lamp and incongruous shadows and reflections from the glass ball are notable, detracting from the authenticity of the scene's ambiance.
After composition, reconfigured objects are adeptly integrated, achieving a coherent and consistent global scene. The changes in the materials of the lamp, the nightstand, and their reflective surfaces 
demonstrate the system's adaptability to diverse source inputs, which also involves a challenging text prompt containing subtle interplay within a multi-object context. 
In a nutshell, our method unleashes its potential by composing scenes with reusable NeRF components, which also facilitates the following editing for scenes described by lengthy sentences even with complex relationships among objects.

\begin{figure}[t!]
    \centering
    \includegraphics[width=\linewidth]{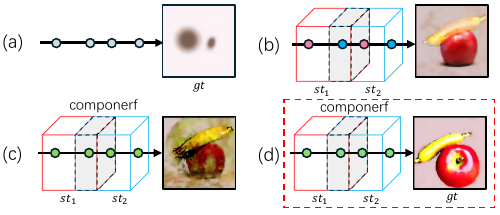}
    \caption{\textbf{The composition strategy.} Our proposed strategies for multi-object scene composition align with \cref{eq:multi_volrend}. The areas of NeRF overlap are indicated in gray. The green nodes represent composited samples. Our design is highlighted by the \underline{dashed box}.
    }
    \label{fig:abls_compo}
\end{figure}

\begin{figure}[!t]
    \centering
    \includegraphics[width=\linewidth]{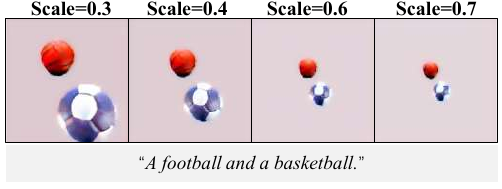}
    \caption{\textbf{Effect of adjusting camera distances.} Scaling camera distance can benefit rendering quality.
    }
    \label{fig:dis_pose}
\end{figure}
\subsection{Ablation Study}
\noindent \textbf{Composition Strategy.}
Our initial tactic, as shown in ~\cref{fig:abls_compo}(a), attempts to manage this by assigning each object to a distinct NeRF with a shared context, diverging from the traditional single-network scene encodings~\cite{poole2022dreamfusion,lin2022magic3d,metzer2022latent,wang2022score}. 
This approach aims to improve object recognition, yet it's prone to ‘guidance collapse' due to the global text prompts' inability to offer precise semantic delineation for individual objects.
Inspired by more accurate rendering of single objects with targeted subtext prompts as seen in \cref{fig:intro}, the subsequent refinement in (b) utilizes specific textual supervision for each object. This tailored guidance confirms that diffusion models more effectively render individual objects when provided with explicit textual for each, thereby overcoming the ‘guidance collapse’ issue in complex multi-object scenarios.
%
Nevertheless, the direct rendering of NeRFs often results in incorrect occlusions within their overlapping regions, indicating a deficiency in global refinement.
To improve it, as detailed in (d), we refine sampling points that were originally guided by texts corresponding to individual objects, now with the incorporation of global textual guidance. This supplementary global oversight guarantees a harmonious rendering that upholds the unique identities of the objects while fostering overall compositional unity. Its vital role is underscored by its absence in (c), where its omission brings collapsed results. We refer readers to our \textit{suppl.} for more ablation studies on this composition strategy. 

\begin{figure*}[t!]
    \centering
    \includegraphics[width=0.8\linewidth]{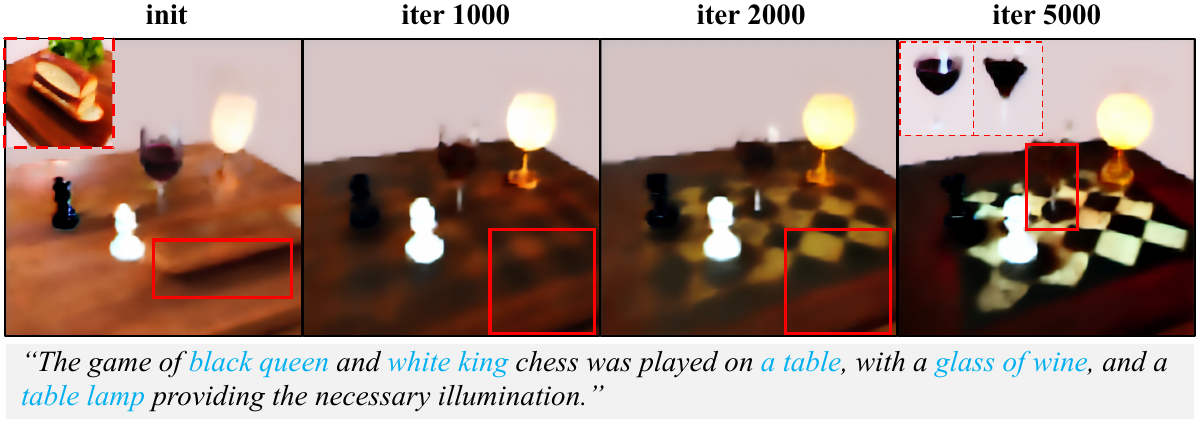}
    \vspace{-15pt}
    \caption{
    \textbf{Recomposition process}. Individual components are labeled in \textcolor{blue}{blue}, while \textcolor{red}{red} boxes emphasize areas of contrast.
    }
    \label{fig:obs}
\end{figure*}
\noindent \textbf{Optimizing Diffusion Model Guidance for Scene Resolution.}
A critical aspect is the need to strike a balance between the overall scene's resolution and the rendering details. For example, when a single object is placed within a vast scene, its rendering results may not be clear as it occupying only a few pixels. This small pixel footprint can limit the amount of gradient information received during backpropagation.
\cref{fig:dis_pose} illustrates this scenario using the same text prompt but with varying scales of the global frames, ranging from 0.3 to 0.7. The results underscore a key insight: the more pixel rays an object interacts with, the better the quality of its rendering. This finding is particularly relevant for large-scale scene rendering, where multiple local frames coexist within the same space. A small object in such a setup may receive minimal ray-box interactions, potentially leading to training inefficiencies or collapse.
It's also important to consider that our scene models are optimized in the latent space of the Stable Diffusion model, which has a feature resolution of 
$64\times 64$. However, we decode these latent color features into RGB images with a resolution of 
$128\times 128$. This discrepancy affects the density of rays throughout the space and, by extension, the number of objects that can be effectively rendered.

\begin{figure}[t]
    \centering
    \vspace{-10pt}    \includegraphics[width=\linewidth]{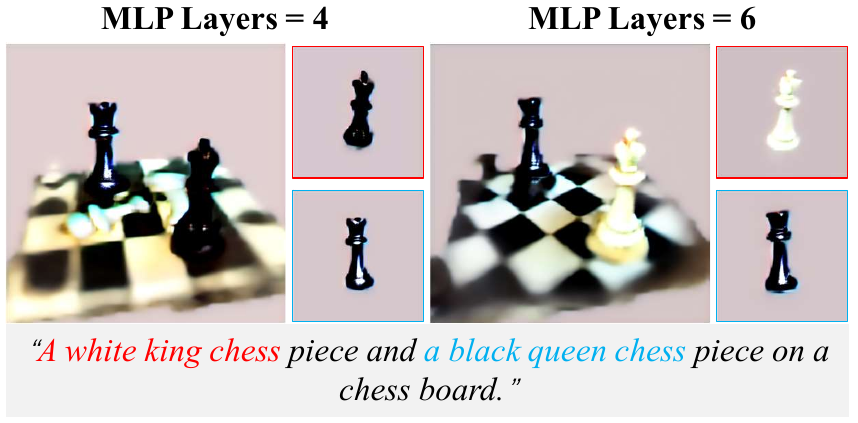}
    \vspace{-18pt}
    \caption{\textbf{Effect of MLP sizes. } Comparison of different parameter sizes for composition. }
    \label{fig:ab_mlp_size}
\end{figure}

\noindent \textbf{Influence of Global MLP Size on Composition Capabilities.}
The complexity of a scene directly influences the required configuration of the parameters \(\boldsymbol{\theta}_g\) within the composition module. In Figure~\ref{fig:ab_mlp_size}, we experiment with varying the number of layers in the MLPs responsible for both densities \(f_{\boldsymbol{\theta}_{g_d}}\) and color calibration \(f_{\boldsymbol{\theta}_{g_c}}\). The results indicate that an insufficiently representative MLP can fail to preserve the distinct identities of individual NeRFs. For instance, the 'white king chess' piece struggles to manifest its characteristic whiteness, leading the global calibrators to compensate inadequately by projecting a flattened representation onto the chessboard surface.
By increasing the number of MLP layers, we observe a notable improvement in the accurate portrayal of each object's identity and overall scene quality. For example, the contextual details of the chessboard, like its grid pattern, are rendered more clearly and naturally.
Based on these findings, we recommend fine-tuning this hyperparameter to align the composition module's capabilities with the specific complexity of the scene.

\noindent \textbf{Global Context Assimilation by Local NeRFs and Composition Module.} 
Despite the primary embedding of the context within the composition module, local NeRFs exhibit an ability to partially learn these global attributes. For instance, \cref{fig:obs} shows that initially, the local NeRF optimization does not occur in isolation; the table from Case 3 in \cref{fig:sota}, for example, bears a residual shadow from the 'French bread' in the original configuration as depicted in the upper left corner.
As the training progresses, these anomalies are resolved, and the local NeRFs gradually assimilate aspects of the global texture, albeit to a limited extent. For instance, while the black and white pattern of the chessboard is predominantly captured by the global composition module, the local representation of the table, highlighted in red box in their upper left corners, remains unchanged. However, as iterations advance,
despite there's no NeRF that is responsible for 'chess board', 
the global frame begins to discern it as the necessary environment and replicate the underlying 'chessboard' pattern. 
This reveals that NeRFs can initially embed global environmental context, while the composition module can possibly merge some necessary local patterns for consistency. 
On the other hand, the early stop may benefit 
the potential degradation in specific elements, such as the 'wine', which worsens as training progresses, as evidenced by the local frames' comparison in the upper left corner. Concurrently, the global rendering depicts the 'wine' as nearly imperceptible. This deterioration hints at the possibility that continued optimization may inadvertently diminish the representation of certain objects.
%


\section{User Study}

In order to comprehensively evaluate the quality of the quality of 3D assets generated using our method, we conducted a user study involving 15 participants. The study consisted of three tasks: a composition correctness task, a generative quality evaluation task, and an object identification task.
The specific design of each task will be elaborated in subsequent sections. Data collection involved both quantitative analysis, through subjective ratings, and qualitative research, carried out via semi-structured interviews.

\subsection{Participants}
\begin{figure}[t!]
    \centering
    \includegraphics[width=\linewidth]{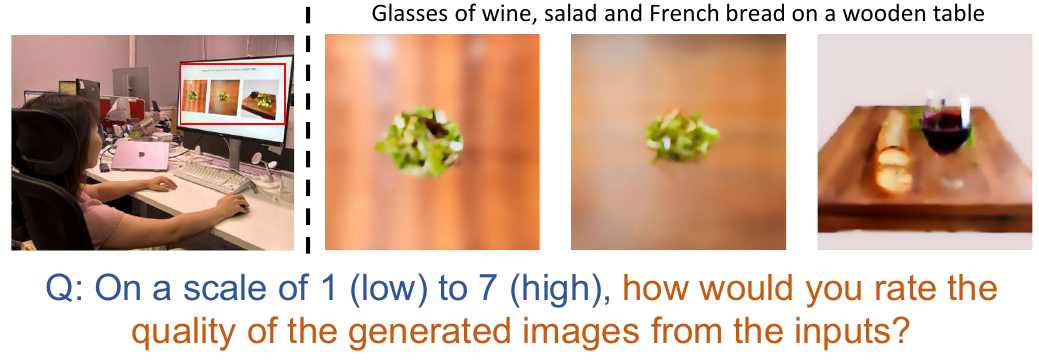}
    \vspace{-15pt}
    \caption{\textbf{User study with our questions demo.}
    }
    \label{fig:user}
\end{figure}

The study included 11 participants, with 54.5\% falling within the 18-24 age group and 45.5\% in the 25-34 age range. The gender distribution was 63.6\% male and 36.4\% female, reflecting a balanced representation. In terms of prior experience, 18.2\% of participants had some familiarity with 3D generative models, while having experience with Blender within the last six months. As depicted in Figure~\ref{fig:user}, the experimental procedure involved participants viewing videos showcasing the 3D assets generated by the model. This approach allowed participants to evaluate the outputs in a controlled and standardized manner, ensuring consistent exposure to the same visual stimuli across all participants.

\subsection{Task design}
\begin{itemize}


\item \textbf{Composition Correctness Evaluation.} In this task, we assess the consistency of the generated 3D assets across two views, focusing on both semantic consistency and multi-view consistency. We collected four groups of samples, each comprising 3D assets generated by Latent-NeRF, SJC, and our method, using the same text prompt. Participants were asked to evaluate the generative consistency from both semantic and multi-view perspectives on a scale from 1 (low) to 7 (high).

\item \textbf{Generative Quality Evaluation.} We provided four groups of generated 3D assets (refer to the supplementary material) to each participant. For each group, the 3D assets were created using Latent-NeRF, SJC, and our method, all based on the same text prompt. Participants were then asked to assess the quality of these generated assets. Finally, we evaluated our method's capability for multi-object generation and combination by calculating the match rate between the objects identified by the participants and those described in the prompt.

\item \textbf{Object Identification.} For this task, we selected four samples of 3D assets generated using our method, comprising a total of seven objects. Participants were then asked to identify the objects depicted in these assets. To evaluate the multi-object generation and combination capabilities of our approach, we calculated the accuracy of the participants' object recognition.

\end{itemize}

\subsection{Measurements}
In the first task, we utilized a 7-point Likert scale to measure participants' perceptions across two dimensions, including semantic consistency and multiview consistency. For the second task, we also used the 7-point Likert scale to evaluate the generative quality and made a comparison with existing works, including Latent-NeRF and SJC. Lastly, we evaluate the effectiveness of our method for multi-object combinations by counting the accuracy of subjects in finding the objects provided in the prompt from the generated 3d assets. We also encourage participants to provide feedback on their emotional responses
and immersive experiences during the overall experience. For details of questions, please refer to \textit{suppl.}

\begin{figure}[t!]
    \centering
    \includegraphics[width=\linewidth]{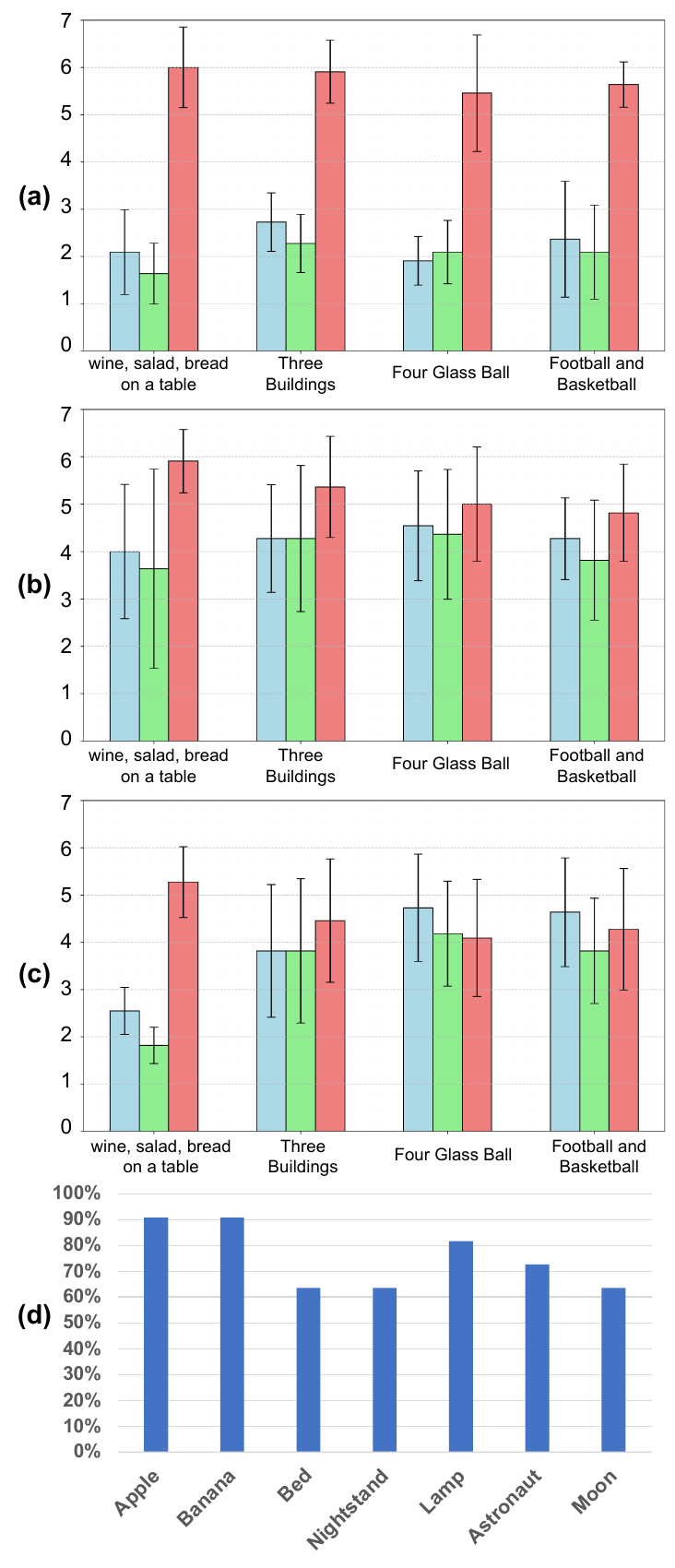}
    \vspace{-10pt}
    \caption{(a) Generative Consistency Evaluation (Semantic); (b) Generative Consistency Evaluation (Multi-view); (c) Generative Quality Evaluation; (d) Object Identification Task.
    }
    \label{fig:user_result}
    \vspace{-8pt}
\end{figure}

\subsection{Results}
As shown in Fig.~\ref{fig:user_result}(a), our method demonstrates significantly improved semantic consistency across all four cases, markedly outperforming previous approaches.
As shown in Fig.~\ref{fig:user_result}(b), benefiting from the reduction of semantic chaos, the multi-view consistency is also better than previous methods.
As shown in Fig.~\ref{fig:user_result}(c), in terms of generative quality evaluation, including the rendering quality of videos and object details, the performance of the three methods is comparable due to their utilization of the same Stable Diffusion version and identical training resolutions.
As shown in Fig.~\ref{fig:user_result}(d), the experiments demonstrate that the majority of our generated 3D scenes are correctly recognized, affirming the semantic consistency between the generated content and the input prompt.

\noindent \textbf{Discussion:} Our observations indicate that users prioritize the level of detail in individual objects when assessing the quality of generated scenes, even if the objects' semantics do not match the given text prompts. For example, as shown at our \textit{suppl.}, when presented with the prompt 'a football and a basketball,' users favored our baseline results, which depicted a single 'football' or 'basketball' with more intricate details, thus appearing more realistic. 
In contrast, our generated scenes, which included more objects, were perceived as less clear due to the increased complexity in rendering.

\section{Conclusion and Future Work}
In this work, we have proposed 
a novel framework for multi-object text-guided compositional 3D scene generation with an editable 3D scene layout.
Our framework interpreted a multi-object text prompt as a collection of localized NeRFs, each associated with a spatial box and an object-specific text prompt, which were then composited to render the entire scene view. We have further enhanced the framework with a specialized composition module for global consistency, effectively mitigating the issue of guidance collapse in the multi-object generation. Utilizing Stable Diffusion model, we have demonstrated that our method, the first to apply a compositional NeRF design to the text-to-3D task, can produce high quality 3D models that feature multiple objects and perform well compared with contemporaneous methods.
Looking ahead, we have explored a promising application of CompoNeRF in the realm of scene editing, which allows for the reuse of trained models in scene recomposition. This capability opens up new possibilities and identifies a rich vein of future work to be pursued in this domain.



\bibliographystyle{abbrv-doi}

\bibliography{template}
\clearpage






\firstsection{Supplementary Material}
\label{sec:sup}
We provide more details of the proposed method and experimental results in the
supplementary material.
Sec.1 and Sec.2 provide the algorithm and more implementation details.
Sec.3 provides more insights into our CompoNeRF model.
Sec.4 adds more details of visualization results. 
Sec.5 lists our attached material details for both scene reconstruction and editing.
\section{Algorithm}
\label{sec:algo}
The detailed algorithm of training our proposed
CompoNeRF is shown in Algorithm ~\ref{alg:cap}.

\begin{algorithm*}[tb]
\caption{Training for CompoNeRF}\label{alg:cap}
\textbf{Input}: a pre-trained text-to-image diffusion model $\phi$, multi-object text prompt $T$ and a set of boxes for 3D scene layout.
\textbf{Output}: learned parameters of local NeRFs $\{\theta_{l,i}\}_{i=1}^{m}$ and Global MLP $\theta_g$.
\begin{algorithmic}[1]
\FOR{\text{\textbf{Iter} $=0 <$ \text{\textbf{MaxIter}}} }
\STATE  Sample $H\times W$ rays from the random camera position and add the directional prompt into $T$.
\FOR{$i=0~\text{to}~H\times W$}

\STATE  Calculate the ray-box intersection for ray $r_i$ to get $m_i$ hits.

\FOR{$j=0 $ to $m_i$}{
    \STATE  Sample $N$ points with normalized location in the $j$ hit  local frame.
    \STATE  Calculate color $\boldsymbol{C_l}$ and density $\boldsymbol{\sigma}_l$ for each point from $\theta_{l,j}$.
    \STATE  Calculate the volumetric rendering color $\hat{\boldsymbol{C_{l,i,j}}}$ of the local Frame for ray $r_i$.}
\ENDFOR
\ENDFOR
\STATE  Map all points into global locations and sort them according to the depth.
\STATE  Calculate the calibrated color $\boldsymbol{C}_{g,i}, \boldsymbol{\sigma}_{g,i}$ via Eq.~4 and Eq.~5 for each point.
\STATE  Calculate the global volumetric rendering color $\hat{\boldsymbol{C_{g,i}}}$ for each ray $r_i$.
\ENDFOR
\STATE  Generate the local view from $\{\hat{\boldsymbol{C_{l,i,j}}}\}_{i=1}^{H\times W}$ and the global view from $\{\hat{\boldsymbol{C_{g,i}}}\}_{i=1}^{H\times W}$
\STATE  Perform score distillation sampling on the local render view and the global render view.
\STATE  Update network parameters via an Adam optimizer.
\end{algorithmic} \textbf{Eng}: Decompose local NeRFs and cache them into offline dataset.
\end{algorithm*}
\section{Implementation Details}
\label{sec:imple}
For score distillation sampling, we use the v1-4 checkpoint of Stable Diffusion based on the latent diffusion model~\cite{rombach2022high}. 
We utilize the code-base~\cite{metzer2022latent} for 3D representation and grid encoder from Instant-NGP~\cite{muller2022instant} as our NeRF model. The global MLP consists of 4 or 6 Linear layers with 64 hidden channels.
In the training loss, we set $\alpha_g=100, \alpha_l=100$, and $\beta=5e^{-4}$ if without specification.
Our 3D scenes are optimized with a batch size of 1 using the Adam~\cite{kingma2014adam} optimizer on a single RTX3090.
Our global frame is centered at the world origin and has a normalized side length of [-1,1]. To generate camera positions, we uniformly sample points on a hemisphere that covers the global frame with a random radius between 1.0 and 1.5.
The camera distance can also be scaled in the way discussed in the main paper. Plus,  
cameras are oriented to look toward the objects. During optimization, the camera field of view is randomly sampled between 40 and 70 degrees. At test time, the field of view is fixed at 60 degrees.
We use the Adam optimizer and perform gradient descent at a learning rate of 0.001 for 5,000 steps in simple prompts, such as ``apple and banana``, and 8,000 steps in more complex prompts for better quality. 
We follow the implementation of SJC~\cite{wang2022score} to perform the averaging implicitly, relying on the optimizer's momentum state when applying the perturb-and-average scoring strategy during training.

\begin{figure}[t]
    \centering
    \includegraphics[width=\linewidth]{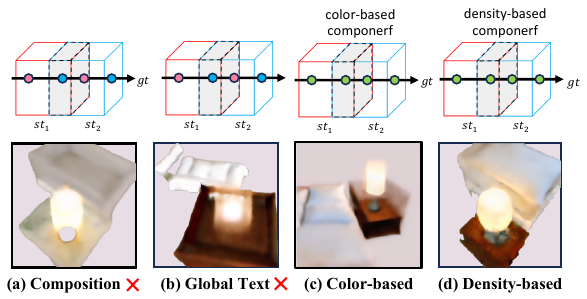}
    \caption{Ablation study on module designs with the scene bedroom. (a) without global calibration. (b) without global text loss. (c) color-based design. (d) our density-based design.}
    \label{fig:sup_module}
\end{figure} 
\begin{figure}[t]
    \centering
    \includegraphics[width=\linewidth]{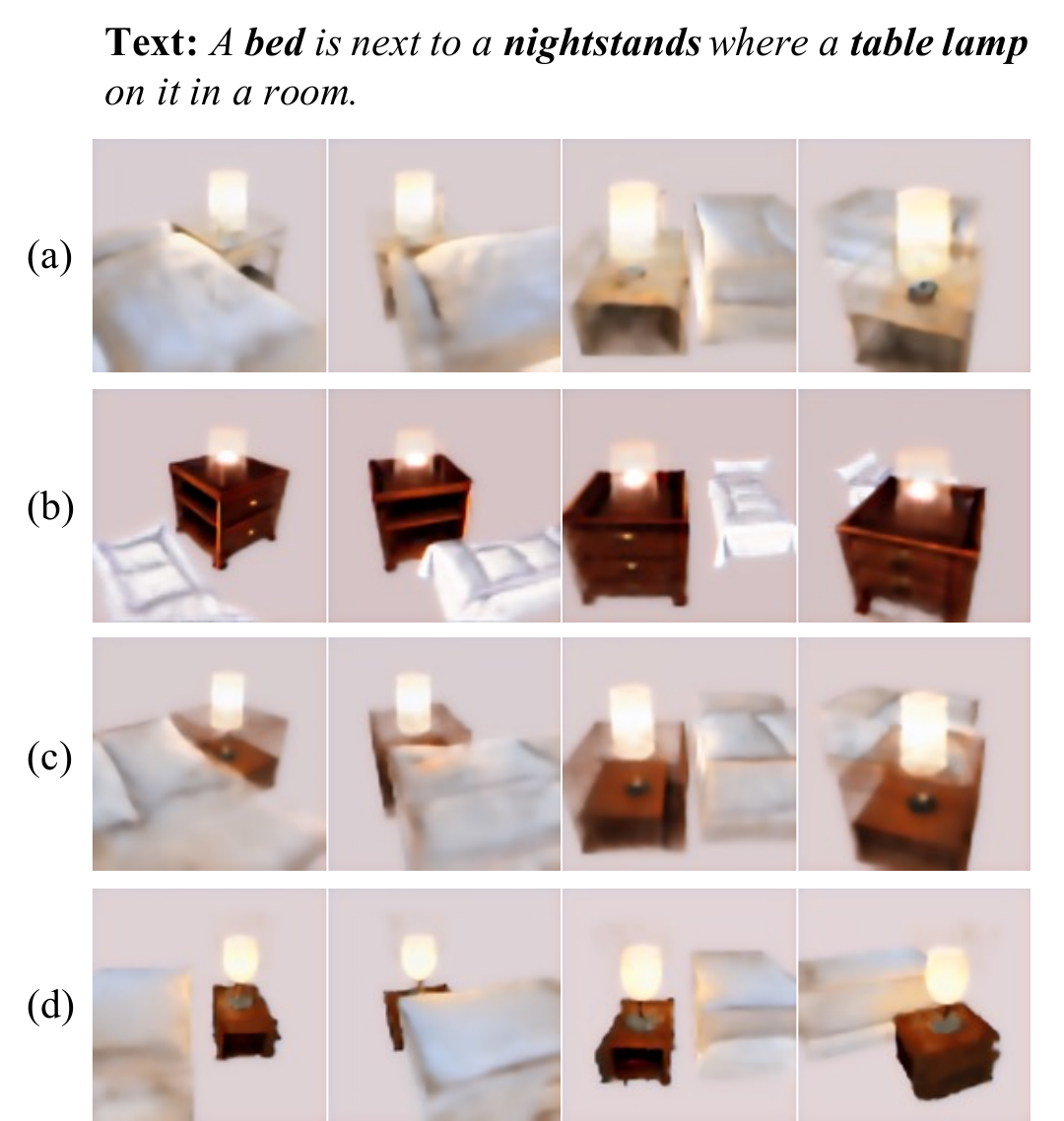}
    \caption{Multi view results with the text prompts on \cref{fig:sup_module}. Subtexts for individual NeRFs are highlighted in bold. }
    \label{fig:ab_view}
\end{figure}

\section{Discussion}


\noindent \textbf{More ablations on the design of composition module.}
In \cref{fig:sup_module}, we present further results from the ablation study of our composition module. As outlined in our main manuscript, our preference for a density-based approach is due to its effective and precise calibration of global density.
For example, the 'bedroom' scene builds upon the discussion from Fig.2(b.2) in the main paper. The complementary study in \cref{fig:sup_module}(a) demonstrates that direct global text supervision without compositional integration leads to a loss of material context, washing out the 'bed' and 'nightstand' in white. Conversely, \cref{fig:sup_module}(b) illustrates that omitting global text and relying solely on subtext supervision retains the familiar context of a 'white sheet' bed and a polished tan 'nightstand'. However, this approach introduces geometric inconsistencies, such as an overly tall nightstand and a lamp lacking a base, along with an absence of light reflection in the surrounding space.
The application of our composition module, depicted in \cref{fig:sup_module}(c) and 8(d), reveals that the density-based design affords enhanced control over density and, consequently, finer geometry. Instead of an empty space above the nightstand, the design aims to adjust the nightstand's height to achieve scene harmony, although it still exhibits limitations in controlling density, leading to subdued floating radiance.
\cref{fig:ab_view} provides a comprehensive visual comparison within the 'bedroom' context. When global calibration is absent, as seen in \cref{fig:ab_view}(a), the scene is plagued by sparse holes and a loss of color and texture detail. Neglecting the global branch entirely, as shown in \cref{fig:ab_view}(b), results in a lack of global consistency, evident in the disproportionate size of the 'nightstand' relative to the scene. Finally, the color-based solution in \cref{fig:ab_view}(c) fails to effectively correct the geometry, introducing additional artifacts. In contrast, the full model in \cref{fig:ab_view}(d) exhibits a marked improvement in these aspects.
\begin{figure}[t!]
    \centering
    \includegraphics[width=\linewidth]{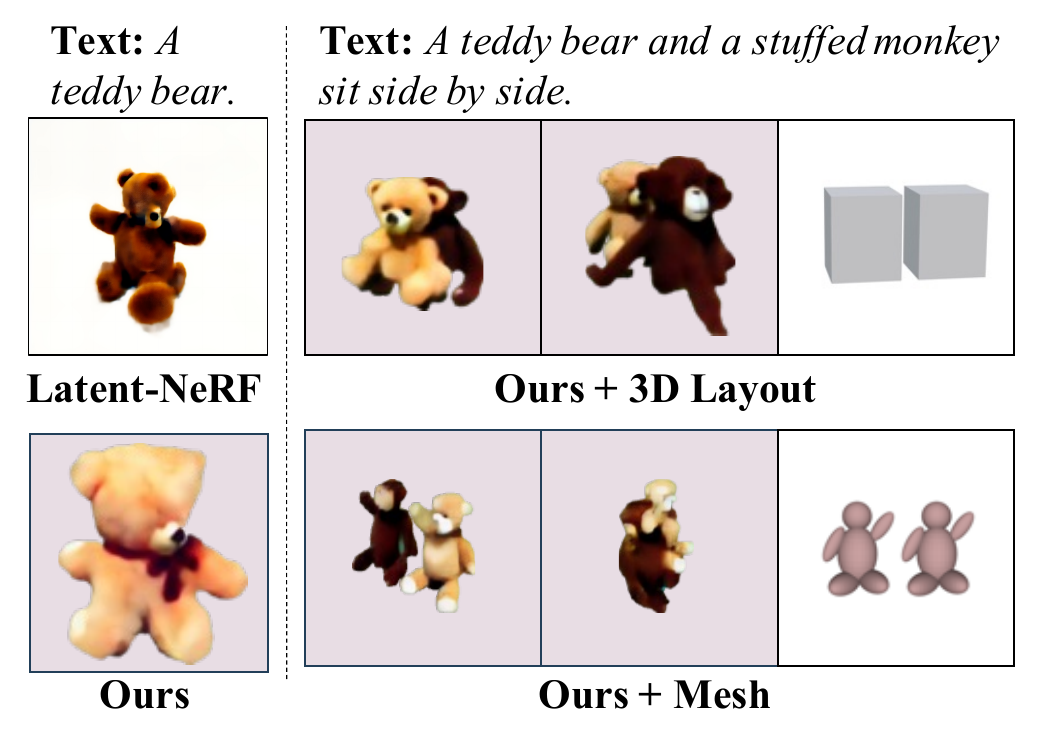}
    \caption{\textbf{(Left)} we observe the multi-face problem, \textit{e.g.}, duplicated face views with geometry collapse in all methods, even in single-object cases. \textbf{(Right)} We provide mesh as guidance instead of box layouts to solve this problem, which further proves our method's versatility and effectiveness.}
    \label{fig:dis_shape}
\end{figure}

\begin{figure}[t!]
    \centering
    \includegraphics[width=\linewidth]{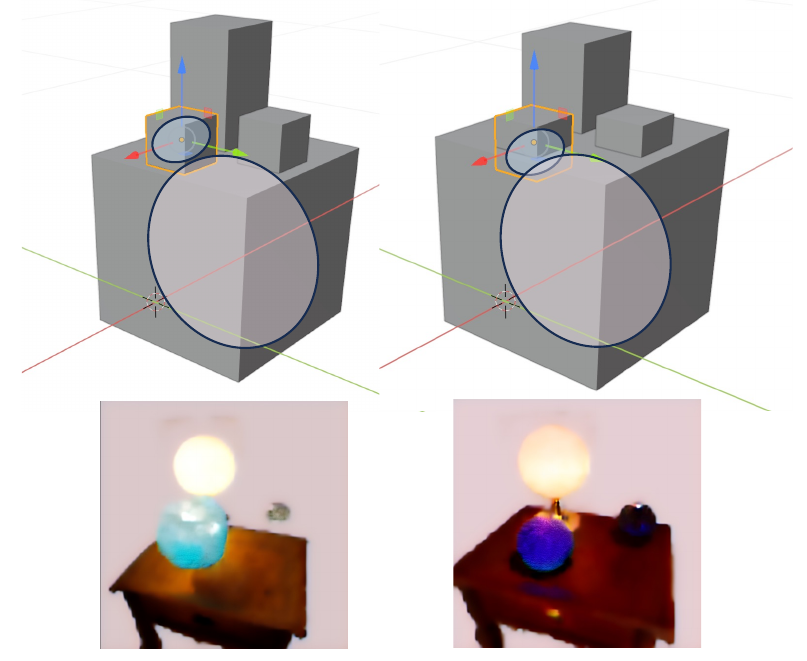}
    \caption{
An ablation study examines layout editing and the floating issue. The upper row shows the layout editing and the lower row indicates rendering views.\textbf{(Left)} Renderings exhibit floating objects due to a suboptimal layout. \textbf{(Right)} Improved outcomes following layout refinements.}
    \label{fig:layout}
\end{figure}
\noindent \textbf{Addressing the multi-face issue with enhanced prompts.}
Much like Latent-NeRF and SJC, our CompoNeRF framework encounters the multi-face challenge, where guidance from the Stable Diffusion model may result in conflicting facial features for certain objects, as illustrated in Figure~\ref{fig:dis_shape}. The reason lies in the fact that diffusion model does not always provide reliable guidance that aligns with the desired orientation corresponding to the camera's viewpoint during sampling.
To mitigate the multi-face problem, stronger constraints can be introduced to promote geometric consistency within the 3D representation. CompoNeRF incorporates mesh constraints, akin to those utilized in Latent-NeRF, offering a more detailed 3D layout compared to traditional bounding boxes. As demonstrated in Figure~\ref{fig:dis_shape}, the implementation of exact mesh constraints markedly mitigates the multi-face issue, though it may come at the expense of detail and adaptability.
Nevertheless, the requirement for accurate mesh input necessitates considerable manual editing, which may reduce the method's range of applications. Despite this, our approach illustrates that the 3D scene layout can be readily adapted to accommodate a broader range of input prompts. Further study is needed to solve the persistent multi-face issue in the text-to-3D tasks.

\noindent\textbf{Adjusting Layout to address Floating Artifacts.}
The process of scene composition begins with strategically positioning NeRFs within the predefined layout.
An overlap of object bounding boxes is critical, as highlighted in Fig.2 of the main document, to facilitate the generation of convincing scenes.
In our investigations, demonstrated in \cref{fig:layout}, we identified a 'floating' issue when bounding box overlaps are absent. This issue may stem from the regularization behavior within NeRFs, where the radiance fields—specifically, the regions responsive to gradient interactions, symbolized by ellipses centered on the boxes—fail to intersect. Such non-interaction can pose challenges, as it does not provide the necessary contiguous context for the global semantics to incorporate these objects seamlessly.
To rectify this, one straightforward approach we recommend is the judicious repositioning of bounding boxes to introduce overlaps. For example, a slight downward adjustment of a box can instigate detectable overlaps during training, facilitating better integration.
This insight opens up a potential avenue of research into the interplay between layout configurations and NeRFs, offering the possibility of more nuanced control over scene dynamics without the need for explicit layout modifications.
\begin{table*}[t!]
\centering
\resizebox{\linewidth}{!}{ 
    \begin{tabular}{l|c|c|c|c|c|c}
    \toprule
    \textbf{Methods} & \textbf{Diffusion Model} & \textbf{\textbf{3D Representation}} & \textbf{Scene Rendering} & \textbf{Input Prompt} & \textbf{Scene Editing} & \textbf{recomposition}\\  \hlineB{2.5}
    DreamFusion & Imagen & Mip-NeRF 360~\cite{Barron2021MipNeRF3U} & Object-centric & Text & T & \XSolidBrush\\    \hline
    Magic3D & eDiff-I + SD & Instant-NGP~\cite{Mller2022InstantNG} & Object-centric & Text & T & \XSolidBrush\\    \hline
    DreamBooth3D & DreamBooth+DreamFusion & Mip-NeRF~\cite{Barron2021MipNeRFAM} & Object-centric & Text+Images & T & \XSolidBrush \\    \hline
    Points-to-3D & ControlNet+Point-E & Instant-NGP & Object-centric & Text+Image & T & \XSolidBrush \\  \hline
    Fantasia3D & SD+PBR & DMTET~\cite{Shen2021DeepMT} & Object-centric & Text/Fine Shape & T/M/S/R & \Checkmark \\
    \hline
    Latent-NeRF  & SD V1.4 & Instant-NGP & Object-centric & Text+Fine Shape & T & \XSolidBrush \\    \hline
    SJC~\cite{wang2022score} & SD V1.5 & voxel radiance field & Object-centric & Text & T & \XSolidBrush\\  \hline
    \hline
    Set-the-Scene~\cite{Cohen-Bar_Richardson_Metzer_Giryes_Cohen-Or_2023} & SD V2.0 & - & Object-compositional & Text+3D Layout & T/M/S/R & \Checkmark  \\ 
    \hline
    \textbf{Ours} & SD V1.4 & Instant-NGP & Object-compositional & Text+3D Layout & T/M/S/R & \Checkmark \\  
    \bottomrule
     \hlineB{1.5}
    \end{tabular}
}
\caption{\textbf{Comparison of our method with the related works for text-to-image generation}. SD denotes Stable Diffusion. For scene editing, we use T(editing object with text), M(moving object), S(scaling object), and R(removing object) for short.}
\label{tab:compare}
\end{table*}
\begin{figure*}[t]
    \centering
    \includegraphics[width=\linewidth]{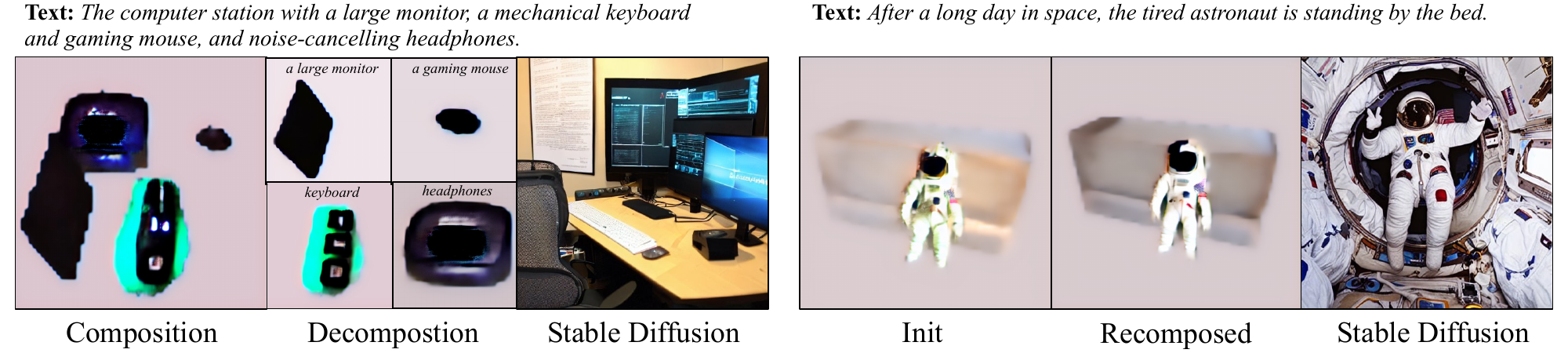}
    \caption{The failure cases are categorized as follows: \textbf{(Left)} issues encountered during scene reconstruction, and \textbf{(Right)} challenges arising in scene editing. The outputs from Stable Diffusion selected for illustration represent the most frequently occurring types generated by the model.
 }
    \label{fig:fail}
\end{figure*}
\begin{figure*}[t!]
    \centering
    \includegraphics[width=\linewidth]{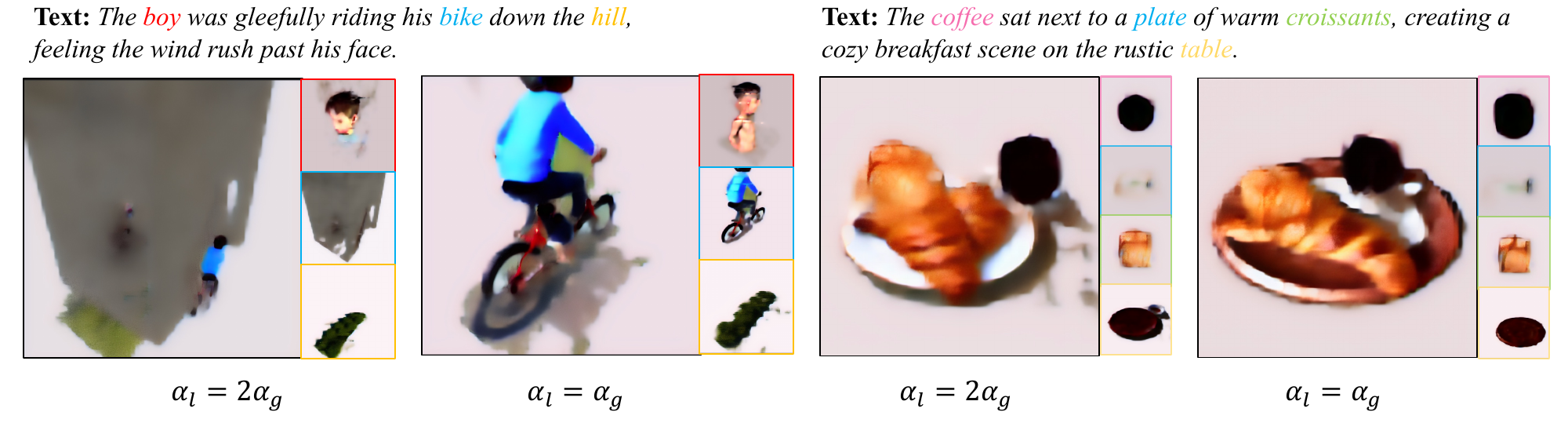}
    \caption{
    The failure cases with their color-labeled NeRFs components are shown beside. The textual guidance manipulation on global/local weights is shown below.  
    }
\label{fig:sup_semantic}
\end{figure*}

\noindent \textbf{Analysis of Failure Cases in Scene Composition and Editing.}
Our composition module may sometimes fail to produce coherent scenes, often due to limited text description distributions within the training data of diffusion models as illustrated in \cref{fig:fail}. This can be mitigated by adjusting the loss weights governing the global and local guidance, such as \cref{fig:sup_semantic}.
In scene composition, the 'computer station' lacks accessories like cables and wires, the 'headphones' are misshapen, and the 'computer screen' lacks a base. Scene recomposition similarly shows the 'astronaut' and 'bed' placed together without sensible global calibration.
Moreover, the Stable Diffusion model's depiction of human figures often suffers from geometric distortions, potentially due to the multi-face problem, as shown in \cref{fig:sup_semantic}.
These failures are mostly due to uncommon layouts or the rarity of certain objects in text-to-image datasets. Repeated global text prompts on Stable Diffusion and examination of numerous samples have failed to yield images that align with our objectives. This challenge extends beyond guidance collapse, reflecting the scarcity of certain objects in the model's outputs.
CompoNeRF's effectiveness is inherently linked to the performance of large-scale text-to-image models, restricting its capabilities to generating primarily conventional scenes with well-defined global features.
To address these limitations, we can strategically adjust the weights of global textual guidance \({\alpha_g}\nabla\mathcal{L}_{\text{SDS}_g}\) and local textual guidance \({\alpha_l}\nabla\mathcal{L}_{\text{SDS}_l}\). This adjustment aims to find an equilibrium between the consistency of the overall scene and the accuracy of individual components. For example, increasing \({\alpha_g}\) enhances global consistency, as evidenced in Figure~\ref{fig:sup_semantic}. However, this can inadvertently lead to objects assimilating extraneous global context. In the 'boy riding bike' scenario, a heightened \({\alpha_g}\) may result in the 'bike' being erroneously represented as both a human figure and a bike. Similarly, in the 'breakfast' scene, amplifying global context might result in a more proportionate table, yet it complicates the distinction between the 'croissant' and its individual NeRF representation.

Ultimately, fine-tuning loss weight parameters is a delicate process that can mitigate identity issues, yet it demands careful calibration to maintain a harmony between scene integrity and the authenticity of each component. The limited representation of certain objects in pre-trained models remains a substantial obstacle, underscoring the need for further investigation into the issue of inadequate guidance in complex scene generation.

\begin{figure*}[t]
    \centering
    \includegraphics[width=0.8\linewidth]{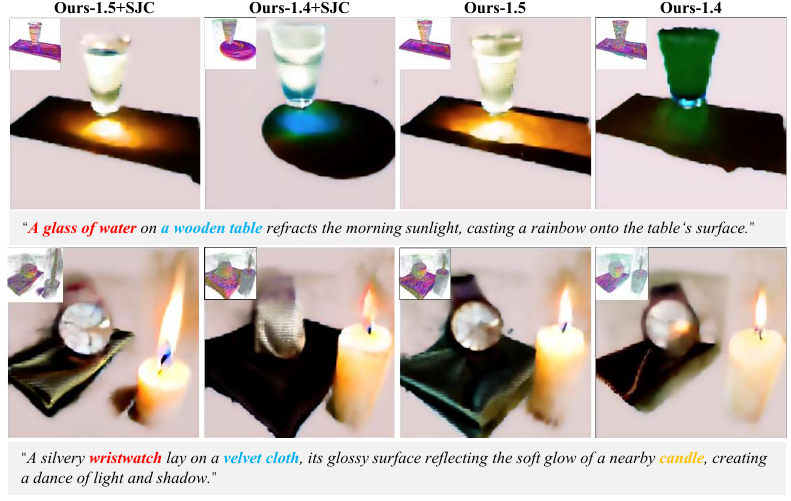}
    \caption{
    The influence of different weight versions of Stable Diffusion.
    }
    \label{fig: abls_version}
\end{figure*}

\noindent \textbf{Influence of SD version.}
As depicted in Fig.~\ref{fig: abls_version}, there is a significant gap between the results using SD V1.5 and SD V1.4, suggesting that upgrading to a more advanced diffusion model could further enhance the quality of the generated content.



\section{More Visualization Results}
\label{sec:vis}
\begin{figure*}[t]
    \centering
    \includegraphics[width=.98\linewidth]{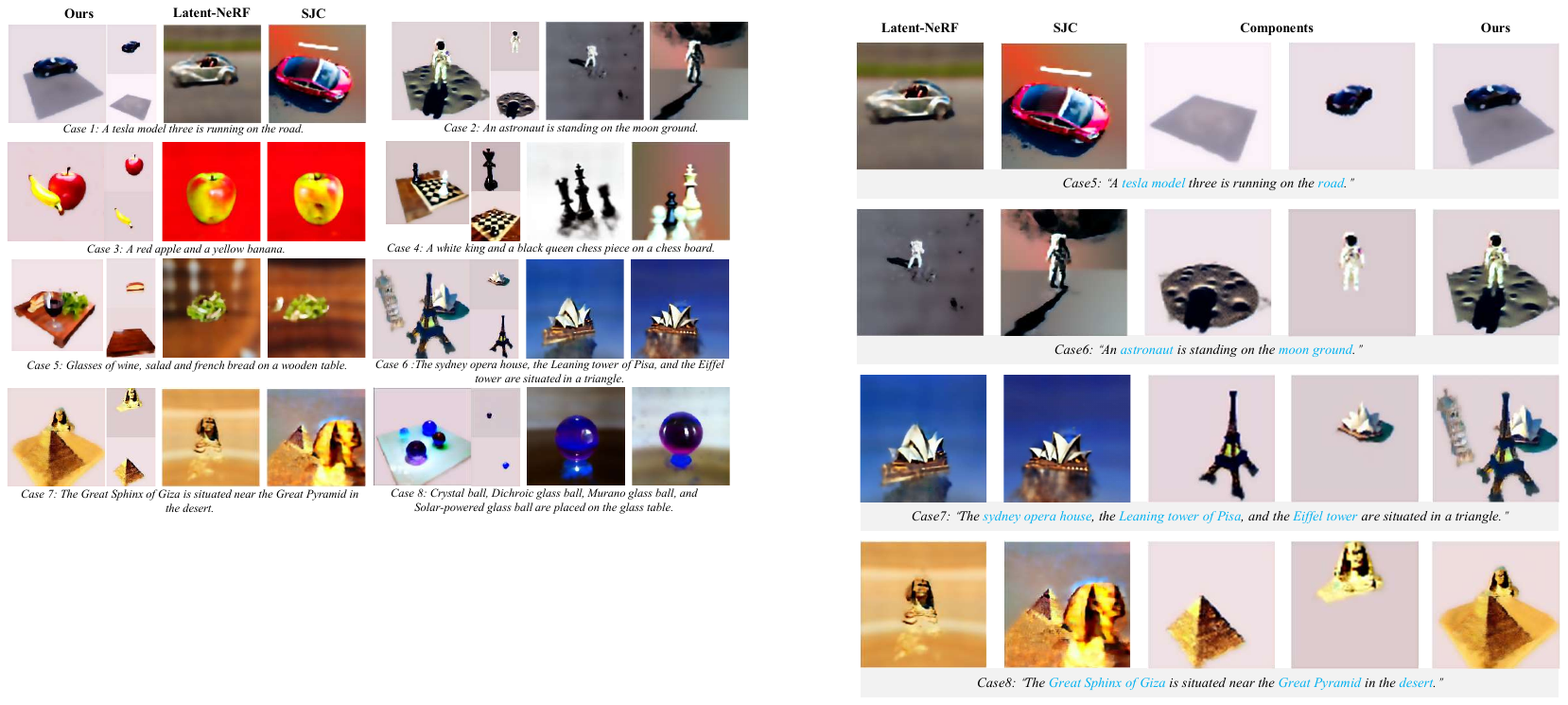}
    \caption{
    More qualitative results using multi-object text prompts. 
    }
    \label{fig:vis1}
\end{figure*}

We provide the multi-view qualitative results from CompoNeRF in \cref{fig:vis1}. 
Note that we increase the resolution of the image latent features from $64 \times 64$ to $128\times128$ during inference for better results.
We also have attached video results in the supplemental materials for each case and the baseline Latent-NeRF~\cite{metzer2022latent} and SJC~\cite{wang2022score}. 
Please see the attached video for rotating frame results. 


\begin{figure*}[t]
    \centering
    \includegraphics[width=.98\linewidth]{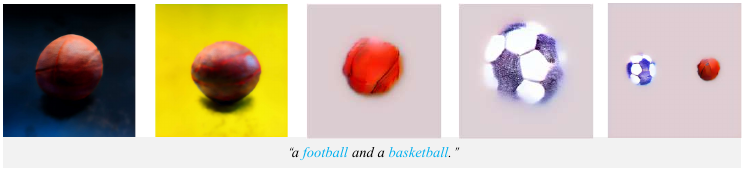}
    \caption{
    More qualitative results using multi-object text prompts. 
    }
    \label{fig: basketball and soccer}
\end{figure*}


\begin{table}[t!]
\renewcommand{\arraystretch}{1.2}
\fontsize{4pt}{4pt}
\selectfont 
\centering
\resizebox{\linewidth}{!}
{
\begin{tabular}{lcccc}
\hline
Method                   & Case 5         & Case 6         & Case 7         & Case 8         \\ 
\hlineB{1.1}
LatentNeRF               & 25.16          & 27.07          & 26.32          & 27.43          \\
SJC                      & 23.55          & 27.84          & 27.41          & 25.62          \\
\textbf{CompoNeRF (Ours)} & \textbf{26.13} & \textbf{32.71} & \textbf{28.44} & \textbf{28.96} \\ \hlineB{1.1}
\end{tabular}
}
\caption{\textbf{More quantitative comparisons}. For our evaluation metric, we utilize the average of CLIP scores~\cite{parmar2023zero,zhang2023sine,wang2023imagen} across different views, which serve to assess the similarity between the generated images and the global text prompt. }
\label{tb:perclass}
\end{table}

\clearpage
\section{Case Details}


\begin{enumerate}
        \item Text prompts : 
        \begin{enumerate}
            \item A tesla model three is running on the road.
            \item An astronaut is standing on the moon ground.
            \item A red apple and a yellow banana.
            \item A white king and a black queen chess piece on a chess board.
            \item Glasses of wine, salad and french bread on a wooden table.
            \item  The sydney opera house, the Leaning tower of Pisa, and the Eiffel tower are situated in a triangle.
            \item The Great Sphinx of Giza is situated near the Great Pyramid in the desert.
            \item Crystal ball, Dichroic glass ball, Murano glass ball, and Solar-powered glass ball are placed on the glass table.
            \item A bed is next to a nightstand with a table lamp on it in a room. 
            \item  A bunch of sunflowers in a barnacle encrusted clay vase. 
            \item A silvery wristwatch lay on a velvet cloth, its glossy surface reflecting the soft glow of a nearby candle, creating a dance of light and shadow.
            \item A glass of water on a wooden table refracts the morning sunlight, casting a rainbow onto the table's surface.
        \end{enumerate}
        \item Scene Editing:
        \begin{enumerate}
            \item 
            A bed is next to a nightstand with a vase of sunflowers on it in a room.
            \item A bed is next to nightstands with table lamps on them in a room.
            \item On the polished surface of the mahogany nightstand, a lamp with a shade of woven silk cast a warm glow over a Murano glass ball and its companion, a shimmering Dichroic glass ball, creating a dance of colors against the darkening twilight.
            \item Glasses of juice, salad, apple, and French bread on a wooden table.
            \item The game of black queen and white king chess was played on a table, with a table lamp providing the necessary illumination.
        \end{enumerate}
        \item Questions for User Study:
        \begin{enumerate}
            \item On a scale of 1 (low) to 7 (high), how would you rate the semantic consistency of the generated 3D assets? (Composition Correctness Evaluation task).
            \item On a scale of 1 (low) to 7 (high), how would you rate the multi-view consistency of the generated 3D assets? (Composition Correctness Evaluation task).
            \item On a scale of 1 (low) to 7 (high), how would you rate the quality of the generated images from the inputs? (Generative Quality Evaluation task).
            \item What objects do you find in the 3D assets? (Object Identification task).
        \end{enumerate}
\end{enumerate}
\end{document}